\documentclass[letterpaper]{article}
\usepackage{aaai}
\usepackage{times}
\usepackage{helvet}
\usepackage{courier}

\usepackage{amsmath}
\usepackage{multirow}
\usepackage{siunitx}
\usepackage{booktabs}
\usepackage{subcaption}
\usepackage{algorithm, algorithmicx}
\usepackage{xcolor}
\usepackage{pgf}
\usepackage{algorithm, algpseudocode}

\begin{document}
% The file aaai.sty is the style file for AAAI Press 
% proceedings, working notes, and technical reports.
%
\title{Automatic deep learning for trend prediction in time series data}
\author{Kouame Kouassi and Deshendran Moodley\\
University of Cape Town and Centre for Artificial Intelligence Research\\
18 University Avenue\\
Rondebosch, Cape Town 7700\\
}
\maketitle
\begin{abstract}
\begin{quote}
    Recently, Deep Neural Network (DNN) algorithms have been explored for predicting trends in time series data. In many real world applications, time series data are captured from dynamic systems. DNN models must provide stable performance when they are updated and retrained as new observations becomes available. In this work we explore the use of automatic machine learning techniques to automate the algorithm selection and hyperparameter optimisation process for trend prediction. We demonstrate how a recent AutoML tool, specifically the HpBandSter framework, can be effectively used to automate DNN model development. Our AutoML experiments found optimal configurations that produced models that compared well against the average performance and stability levels of configurations found during the manual experiments across four data sets.  
\end{quote}
\end{abstract}

\section{Introduction}
% In certain time series prediction applications, segmenting the time series into a sequence of trends and predicting the slope and duration of the next trend is preferred over predicting just the next value in the series \cite{Wang2011,Lin2017}[4][5]. Piecewise linear representation [Keogh2004] or trend lines can provide a better representation for the underlying semantics and dynamics of the generating process of a non-stationary and dynamic time series \cite{Wang2011,Lin2017}. Moreover, trend lines are a more natural representation for predicting change points in the data, which may be more interesting to decision makings. For example, suppose a share price in the stock market is currently rising. A trader in the stock market would ask “How long will it take and at what price will the share price peak and when will the price start dropping.” Another example application is for predicting daily household electricity consumption.  Here the user may be more interested in identifying the time, scale and duration of peak energy consumption.\\
In 2017, Lin et al. \cite{Lin2017} proposed one of the first studies that explored the use of Deep Neural Networks (DNN) for trend prediction in times series data. They proposed a hybrid deep neural network algorithm, TreNet, which was shown to have superior performance over other DNN and traditional ML approaches. While TreNet was no doubt an important development in terms of the application of DNNs for trend prediction, the validation method used in the experiments did not take into account the sequential and dynamic nature of most real world time series data sets. 

In recent work \cite{kou2020} we replicated the TreNet experiments on the same data sets using a walk-forward validation and tested our optimal model over multiple independent runs to evaluate model stability. We obtained results that were substantially different from the original TreNet experiments. Moreover, we showed that that simpler and faster vanilla DNN models, i.e. the MLP, LSTM and CNN, performed comparably to the TreNet on the majority of the data sets.
However, the development of these DNN models involved extensive experimentation to find the optimal training and structural hyperparameters. One of the challenges in our study was the time and effort required for manual hyper-parameter optimisation of the experiments.  

In this paper we explore automated machine learning (AutoML) tools for algorithm selection and hyperparameter optimisation for trend prediction. The paper is structured as follows. We first provide a brief background of the problem and a summary of related work, followed by a brief description of the results of the manual DNN experiments. We then present the results of the AutoML experiments and provide compare the results of the manual and automatic model searching experiments. Finally, we provide a summary and discussion of the key findings.
	
\section{Background and related work}

We define a univariate time series as $X = \{x_1, ..., x_T\}$, where $x_t$ is a real-valued observation at time $t$. The trend sequence $T$ for $X$, is denoted by $T = \{<l_1, s_1>, ..., <s_k, l_k>\}$, and is obtained by performing a piecewise linear approximation of $X$ \cite{Keogh2001a}. $l_k$ represents the  \textit{duration} and is given by the number of data points covered by trend \(k\) and $s_k$ is the slope of the trend expressed as an angle between -90 and 90 degrees. Given a historical time series $X$ and its corresponding trend sequence $T$, the aim is to predict the \textit{next trend} $<s_{k + 1}, l_{k + 1}>$.

\subsection{Trend prediction using DNNs}
Traditional trend prediction approaches include Hidden Markov Models (HMM)s \cite{Wang2011,Matsubara2014} and multi-step ahead predictions \cite{Chang2012}. Leveraging the success of CNNs, and LSTMs in computer vision and natural language processing \cite{LeCun1998,Chung2014EmpiricalEO,Guo2016}, Lin et al. \cite{Lin2017} proposed a hybrid DNN approach, TreNet, for trend prediction. TreNet uses a CNN which take in recent point data, and a LSTM which takes in historical trend lines to extract local and global features respectively. These features are then fused to predict the next trend. Lin et al. showed in their experiments that TreNet performed substantially better than other DNN and traditional ML algorithms. However, their study made use of standard cross-validation with random shuffling and a single hold-out set. The use of cross-validation with shuffling implies that data instances, which are generated after the given validation set, are used for training. Besides, the use of a single hold-out set does not provide a sufficiently robust performance measure for data sets that are erratic and non-stationary \cite{Bergmeir2012}. Furthermore, DNNs, as a result of random initialisation and possibly other random parameter settings could yield substantially different results when re-run with the same hyperparameter values on the same data set. In real world applications where systems are often dynamic, DNN models become outdated and must be frequently updated as new data becomes available. It is also crucial that optimal DNN configurations should be stable, i.e. have minimal deviation from the mean test loss across multiple runs. There is no evidence that this was done for TreNet.  Furthermore, many important implementation details in the TreNet study are not stated explicitly. For instance, the segmentation method used to transform the raw time series into trend lines is not apparent. In recent work \cite{kou2020} we compared TreNet and vanilla DNN models across four data sets, three of which was used in the original TreNet paper. We used walk-forward validation and trained and evaluated each configuration across 10 different runs. Our results showed that in general TreNet outperforms the vanilla DNN models but not significantly so. In fact, on one of the datasets, the NYSE dataset, the LSTM with point data features alone outperformed TreNet by 1.29\%.

\subsection{AutoML using BOHB}
Methods for combined algorithm selection and hyperparameter optimisation (CASH) \cite{Feurer2015,Thornton2013}, also referred to as AutoML, can be applied to automate DNN model selection and configuration for trend prediction provided that an appropriate data partitioning and evaluation method is used instead of standard k-fold cross-validation.
Various approaches such as random search \cite{Bergstra:2012:RSH}, evolutionary algorithms for deep learning algorithms \cite{Young:2015:ODL}, Bayesian optimisation (BO) \cite{Thornton2013}, and hyperband (HB) \cite{Li} have been proposed to tackle the CASH problem. BO models the dependency between the hyperparameter configurations and their performance measure using a model. Then, using the current model and an acquisition function, BO predicts the most promising candidate configurations of hyperparameters. Finally, it evaluates the candidate configuration, and updates the model in a Bayesian fashion. The last two steps are repeated until a stopping criterion, such as a computational budget or time limit, is met. Many variants of BO such as SMAC \cite{Thornton2013}, MTBO \cite{NIPS2013_5086}, FABOLAS \cite{Klein2017} have appeared in the literature. BO generally converges to a good configuration but it requires extensive compute resources and/or takes long to gather enough samples to build its model. 

Hyperband (HB) however is a multi-armed bandit strategy that is faster compared to BO. HB takes advantage of the fact that in many applications, the true function $f$ to be learnt can be approximated by a cheap-to-evaluate approximation $\tilde{f}(., b)$, where $b \in [b_{min}, b_{max}]$ is the evaluation budget such as the number of epochs for training DNNs. HB assumes that the quality of the approximation typically increases with the budget $b$. Thus, $\tilde{f}(., b_{max}) = f(.)$. HB exploits this concept to evaluate multiple hyperparameter configurations on cheaper budgets to determine the most promising ones. Then, the promising configurations are evaluated on higher budgets and eventually on the maximum budget to obtain the true function, i.e. the optimal model.HB is faster compared to BO however, it is not as stable as BO because the initial set of configuration is randomly selected. 

Recently, Falkner et al. \cite{Falkner2018} proposed Bayesian Optimisation and Hyperband (BOHB) which combines the strengths of both HB and BO. BOHB makes use of the multi-fidelity evaluation approach of HB but selects the initial set of configurations using a model similar to BO. BOHB can therefore ensure stable and competitive results when limited compute resources are available.

\subsection{AutoML for trend prediction}
Despite the availability of recent automated machine learning (AutoML) frameworks such as BOHB search, there is limited research on the use of AutoML for trend prediction. Most papers on trend prediction select the machine learning algorithm based on expert knowledge and perform extensive or very little experiments to optimise the hyperparameters \cite{Lin2017,Wang2011}. There has been some attempt to use AutoML for other time series problems \cite{Aras2016,Balkin2000,Koki2019,Honda2019}. However, these studies did not focus on trend prediction nor did they incorporate recent AutoML frameworks such as BOHB.

\section{Manual experiments}\label{exp-manual}
This section summarizes the results of experiments carried out on four different time series datasets using a MLP, LSTM and a MLP.  Three of these datasets are the datasets used in the original TreNet paper. Further details about the datasets and experiments using other techniques can be found in \cite{kou2020}.
	
\subsection{Datasets}\label{sec:exp-datasets}
	Experiments were conducted on four different datasets. (1) The \textit{voltage dataset} from the UCI machine learning repository\footnotemark{}. It contains 2075259 data points of a household voltage measurements of one minute interval. (2) The \textit{methane dataset} from the UCI machine learning repository\footnotemark{}. We used a resampled set of size of 41786 at a frequency of 1Hz. (3) The \textit{NYSE dataset} from Yahoo finance\footnotemark{}. It contains 13563 data points of the composite New York Stock Exchange (NYSE) closing price from 31-12-1965 to 15-11-2019. (4) The  \textit{JSE dataset} from Yahoo finance. It contains 3094 data points of the composite Johannesburg Stock Exchange (JSE) closing price from 2007-09-18 to 2019-12-31. 

	\addtocounter{footnote}{-3} %n=6
		\stepcounter{footnote}\footnotetext{
		\resizebox{0.92\columnwidth}{0.8\height}{https://archive.ics.uci.edu/ml/datasets/individual+household+electric+power+consumption}
		}
	\stepcounter{footnote}\footnotetext{
	\resizebox{0.92\columnwidth}{0.8\height}{https://archive.ics.uci.edu/ml/datasets/gas+sensor+array+under+dynamic+gas+mixtures}}
	\stepcounter{footnote}\footnotetext{\resizebox{0.8\width}{0.78\height}{https://finance.yahoo.com}}

	\subsection{Data preprocessing} \label{sec-exp-preproc}
 The segmentation of the time series into trend lines i.e. piecewise linear approximations is done by regression using the bottom-up approach, similar to the approach used by \cite{Wang2011}. The data instances, i.e. the input-output pairs are formed using a sliding window. The input features are the local data points \({L_k} = <x_{t_k-w}, ...,x_{t_{k}}>\) for the current trend $T_k = <s_k, l_k>$ at the current time $t$. The window size $w$ is determined by the duration of the first trend line.  The output is the next trend $T_{k+1} = <s_{k+1}, l_{k+1}>$. 

\subsection{DNN algorithms} \label{sec-exp-algorithm}
	 The configuration of the three DNN algorithms used in the experiments, i.e. the multilayer perceptron (MLP), the long short-term memory recurrent neural network (LSTM-RNN) and the convolutional neural network (CNN), are described below.
\begin{itemize}
    \item \textit{The MLP} consists of $N$ number of fully connected neural network (NN) layers, where, $N \in [1, 5]$. Each layer is followed by a ReLU activation function to capture non-linear patterns. To prevent overfitting, a dropout layer is added after each odd number layer, except the last layer. % For instance, if the number of layers $N = 5$, the layer 1 and layer 3 will be followed by a dropout layer.
	
	\item \textit{The LSTM model} consists of $N$ LSTM layers, where $N \in [1, 3]$. Each layer is followed by a ReLU activation function to extract non-linear patterns, and a dropout layer to prevent overfitting. After the last dropout layer, a fully connected NN layer is added. This layer takes the feature representation extracted by the LSTM layers as its input and predicts the next trend. The LSTM layers are not re-initialised after every epoch.
	
	\item \textit{The CNN model} consists of $N$ 1D-convolutional layer, where $N \in [1, 3]$. Each convolutional layer, which consists of a specified number of filters of a given kernel size, is followed by a ReLU activation function, a pooling layer, and a dropout layer to prevent overfitting. The final layer of the CNN algorithm is a fully connected neural network which takes the features extracted by the convolution, activation, pooling, and dropout operations as its input and predicts the next trend.
\end{itemize}
	The equally weighted average slope and duration mean square error (MSE) is used as a loss function during training using the Adam optimizer \cite{kingma2014adam}. To ensure robustness against random initialisation, the DNNs are initialised using the He initialisation technique \cite{he2015delving} with normal distribution, fan-in mode, and a ReLU activation function.
	
	\subsection{Evaluation and results} \label{sec-evaluation}
	The walk-forward evaluation procedure, with the successive and overlapping training-validation-test partition \cite{LUO2013}, is used to evaluate the performance of the models. The input-output data instances are partitioned into training, validation, and test sets in a successive and overlapping fashion \cite{LUO2013}. We set the number of partitions to 8 for the voltage, 44 for methane, 5 for NYSE and, 101 for the JSE dataset. This determines the number of training-validation-test evaluations for each dataset. For example, 7 (8-1) training-validation-test evaluations are performed for the voltage dataset. 
	For the methane and JSE datasets, the combined test sets make up 10\% of their total data instances as per the original TreNet experiments; and 80\% and 50\% for the voltage and NYSE datasets respectively because of their large sizes. The average root mean square error (RMSE) given in equation~\ref{eq:rmse} is used as the evaluation metric. 
	\begin{equation}
	\label{eq:rmse}
	\resizebox{0.75\width}{!}{$	RMSE = \sqrt{\frac{1}{T}\sum_{t = 1}^{T}(y_t - y_t^{'})^2}$}
	\end{equation} 
	\begin{table}[!htbp]
	    \centering
		\caption{RMSE value achieved by the manually tuned DNN models. The lower the RMSE error the better.}
		\label{tab:vanilla-dl}
		\resizebox{\columnwidth}{!}{
			\begin{tabular} {llllll}
				\toprule
				&  & {\textit{MLP}}  & {\textit{LSTM}}& {\textit{CNN}} & {\textbf{Best DNN}}\\
				\midrule
				\multirow{4}{*}{\textit{Voltage}} 
				& {\textit{S}} & {$\textbf{9.04} \pm \textbf{0.06}$}  & {$10.30 \pm 0.0$} & {$9.24 \pm 0.10$} & \\
				
				& {\textit{D}} & {$62.82 \pm 0.04$} & {${62.87} \pm {0.0}$} & {$\textbf{62.40} \pm \textbf{0.13}$} & {\textbf{CNN}}\\
				& {\textit{A}} & {$35.93 \pm 0.05$} &  {${36.59} \pm {0.0}$} & {$\textbf{35.82} \pm \textbf{0.12}$} &\\
				\midrule
				\multirow{4}{*}{\textit{Methane}} 
				& {\textit{S}} & {$14.57\pm 0.10$} & {$\textbf{14.21} \pm \textbf{0.19}$} & {${15.07} \pm {0.35}$} & \\
				& {\textit{D}} & {$\textbf{49.79} \pm \textbf{4.85}$} & {$56.37 \pm 1.77$} & {$54.79 \pm 4.55$} & {\textbf{MLP}}\\
				& {\textit{A}} & {$\textbf{32.18} \pm \textbf{2.48}$} & {$35.29 \pm 0.49$} & {$34.93 \pm 2.45$} &\\
					
				\midrule
				\multirow{4}{*}{\textit{NYSE}} 
				& {\textit{S}} & {$90.76 \pm 4.43$} & {$\textbf{86.56} \pm \textbf{0.01}$} & {$89.31 \pm 1.38$} & \\
				& {\textit{D}} & {$33.08 \pm 42.08$} & {$\textbf{0.41} \pm \textbf{0.08}$} & {$12.21 \pm 12.17$} & {\textbf{LSTM}}\\
				& {\textit{A}} & {$61.92 \pm 23.26$}  & {$\textbf{43.49} \pm \textbf{0.05}$} & {$50.76 \pm 6.78$} & \\
				
				\midrule
				\multirow{4}{*}{\textit{JSE}} 
				& {\textit{S}} & {${19.87} \pm {0.01}$} & {${\textbf{19.83} }\pm \textbf{0.01}$} & {${19.90} \pm {0.06}$} & \\
				& {\textit{D}} & {$12.51 \pm 0.09$} & {$12.68 \pm 0.01$} & {$\textbf{12.48} \pm \textbf{0.21}$} & {\textbf{MLP/CNN}}\\
				& {\textit{A}} & {$\textbf{16.19} \pm \textbf{0.05}$} & {${16.25} \pm {0.01}$} & {$\textbf{16.19 }\pm \textbf{0.14}$} & \\
				\bottomrule
			\end{tabular}
		}
	\end{table}
	
	Table~\ref{tab:vanilla-dl} shows the average RMSE values for slope and trend predictions achieved by the MLP, LSTM, CNN on each dataset across 10 independent runs. The deviation across the 10 runs is also shown and provides an indication of the stability of the model across the runs. The equally weighted average slope and duration RMSE values are used as an overall comparison metric.

	\section{Automatic model selection}
	Finding the optimal model for a particular time series requires extensive experimentation by a machine learning expert, and often requires information about the characteristics of that time series. We conducted experiments to explore the use of BOHB to automate the manual algorithm selection and hyper-parameter optimisation. We conducted two experiments, the first searches and evaluates configurations across all three algorithms (BOHB-All). In the second we perform individual experiments for each algorithm (BOHB-Single).
	
	\subsection{AutoML framework and implementation}
	We implemented BOHB using the HpBandSter\footnotemark{} framework. We chose BOHB for two main reasons. Firstly, it fulfills five main desiderata: a strong anytime performance; a strong final performance; scalability; and robustness \& flexibility \cite{Falkner2018}. Secondly, it is suitable for applications with low computational resources because it leverages the speed of hyperband \cite{Falkner2018,Li2016}.  Hyperband is fast  because it eliminates sub-optimal hyperparameter configurations by performing many low fidelity, i.e. cheaper model evaluations on smaller budgets, and fewer high fidelity, i.e. costlier model evaluations on near-optimal configurations. 
    The framework requires the specification of a a \textit{loss metric}, the \textit{hyperparameter configuration search space}, and a stopping criterion, or \textit{budget}. The \textit{loss metric} is the equally weighted average slope and duration RMSE. The other two are described next.
		\addtocounter{footnote}{-1} %n=6
	\stepcounter{footnote}\footnotetext{\resizebox{!}{0.8\height}{https://github.com/automl/HpBandSter}}
\subsection{Hyperparameter configuration space}
Based on our experiences during the manual experiments, we analysed and identified key hyper-parameter variables and ranges that we observed had the most impact on model performance.  We distinguished between 4 types of parameters, that specify the structure of the DNN, training parameters, regularisation and the algorithm type. Common hyperparameters such the learning rate, the batch size, and the dropout rate are shared to reduce the search space. The choice of algorithm is represented as a categorical hyperparameter, which splits the search spaces into sub-spaces which contain the hyperparameters specific to that algorithm. For instance, the kernel size parameter for the CNN is only activated in the search space when CNN is selected. The resultant hyper-parameter search space consisted of 24 different hyperparameters, 22 that are categorical or discretised and 2 that are continuous. There is 1 hyperparameter for algorithm type, i.e. MLP, CNN, or LSTM, 6 structural hyperparameters for the MLP, 4 for the LSTM, 9 for the CNN, and 4 common training and regularisation hyperparameters that are shared accross all algorithms. Given the 2 continuous parameters, the number of possible unique configurations is infinite. A summary of the hyperparameter configuration search space is provided in Table~\ref{tab:bohb-hyperparam}. The full hyperparameter sets/ranges is specified using ConfigSpace \cite{ConfigSpace}, and can be accessed in json format via this link\footnotemark{}.

\begin{table}[!htbp]
\centering
\caption{Summary of the hyperparameters in the configuration space}
\label{tab:bohb-hyperparam}
\resizebox{\columnwidth}{!}{    
\begin{tabular}{lll}
\toprule
\textbf{MLP/CNN/LSTM hyperparameter} &  \textbf{Value type} & \textbf{Type}\\
\midrule
Algorithm & Categorical & Algorithm\\
Number of hidden/CNN/LSTM layers & Discrete & Structure \\
Number of hidden neurons/filters/cells of layer\_i & Discrete & Structure \\
Kernel size of CNN layer\_i & Discrete & Structure\\
Pooling type for CNN layers & Categorical & Structure\\
Pooling size for CNN layers & Discrete & Structure\\
Batch size & Discrete & Training \\
Learning rate & Continuous & Training\\
Dropout rate & Discrete & Regularisation\\
Weight decay & Continuous & Regularisation\\
\bottomrule
\end{tabular}
}
\end{table}
	\addtocounter{footnote}{-1} %n=6
	\stepcounter{footnote}\footnotetext{\resizebox{!}{0.8\height}{https://github.com/h-kouame/configuration-space-of-auto-cash}}
	
	\subsection{BOHB budget and configuration}
We use the \textit{number of training epochs} of the model to estimate its fidelity with respect to the true DNN function to be learnt (see related work on BOHB). Thus, the lowest fidelity model is trained with on the \textit{minimum budget}, i.e. with the minimum number of epochs, and the highest fidelity model is obtained when it is evaluated on the \textit{maximum budget}, i.e. with the maximum number of epochs. We used the maximum number of epochs from the manual experiments to guide the number of epochs required to identify the optimal DNN configuration. Thus, for the single algorithm experiments, we set the maximum budget to the maximum number of epochs found in the manual experiments for that algorithm (see table~\ref{tab:budgets}). When the search space contains all the algorithms, the largest maximum budget across all three single maximum budgets is used for the NYSE, and JSE datasets (see table~\ref{tab:budgets}). For the voltage and methane dataset, we used the a third of this value (see table~\ref{tab:budgets}), since the maximum budget is too high, i.e. 15000. This in fact constrained the BOHB to find optimal models that are faster to evaluate. 
	Having chosen the maximum budget, the minimum budget is determined using equation~\ref{eq:min-budget}, 
	where \\$\eta \rightarrow a \; hyperparameter \; of \; the \; hyberband \; algorithm$ \cite{Li2016,Falkner2018}; 
	and $N \rightarrow$ \textit{the  number  of  medium  budgets  between  the  minimum  budget  and  the  maximum  budget.}
	\begin{equation}
	\label{eq:min-budget}
	\resizebox{\columnwidth}{!}{
		$minimum \: budget = \lfloor \frac{maximum \: budget }{\eta^{N + 1}} \rfloor = \lfloor \frac{maximum \: budget }{3^2} \rfloor$
	}
	\end{equation}
	Following Li. et al's recommendation \cite{Li}, $\eta$ is set to 3; and $N$ to 1. The minimum and maximum budget per dataset are provided in table~\ref{tab:budgets}.
	The number of iterations of BOHB was set to 30. All the other BOHB's parameters are kept to their default values except the \textit{top\_n\_percent} and the \textit{num\_samples}, which are respectively doubled to 30, and halved to 32, following Falkner et al.'s \cite{Falkner2018} guidelines\footnotemark{}. 
	\begin{table}[!htbp]
		\caption{BOHB budget (number of epochs), mimimum budget (min), maximum budget (max) used for each datasets }
		\label{tab:budgets}
		\centering
			\resizebox{0.8\columnwidth}{!}{
		\begin{tabular} {lcccccccc}
			\toprule
			& \multicolumn{2}{c}{NYSE} & \multicolumn{2}{c}{JSE} &  \multicolumn{2}{c}{Voltage} & \multicolumn{2}{c}{Methane} \\
			\cmidrule(lr{.75em}){2-3}
			\cmidrule(lr{.75em}){4-5}
			\cmidrule(lr{.75em}){6-7}
			\cmidrule(lr{.75em}){8-9}
			
			& {min } &{ max } & {min } & {max } & {min } & {max } &{ min } & {max }\\
			
			\midrule
			{MLP} & {55} & {500} & {11} & {100} & {555} & {5000} &{1666} & {15000}\\
			
			{LSTM} & {11} & {100} & {11} & {100} & {111} & {1000} &{1666} & {15000}\\
			
			{CNN} & {11} & {100} & {11} & {100} & {1555} & {15000} &{111} & {1000}\\
			\midrule
			{All} & {55} & {500} & {11} & {100} & {333} & {3000} &{333} & {3000}\\
			
			\bottomrule	
		\end{tabular}
			}
	\end{table}
	\addtocounter{footnote}{-1} %n=6
	\stepcounter{footnote}\footnotetext{\resizebox{0.92\columnwidth}{0.8\height}{https://automl.github.io/HpBandSter/build/html/best\_practices.html}}
	
	\subsection{Number of configurations evaluated}
	The number of unique configurations evaluated for each manual and automatic experiment is shown in table~\ref{tab-cash-manual-configuration}. For the manual experiments each configuration was evaluated 10 times. For the BOHB experiments, a certain number of configurations are evaluated first on the minimum budget (number of epochs). Then, the budget is increased to the medium budget, i.e. $3 \times (min \: budget)$) and the best half configurations are evaluated. Finally, the best half of the medium budget configurations are evaluated with the  maximum budget (highest fidelity). Thus, promising configurations are first identified using a minimum budget and further explored using a higher budget. For the  single algorithm experiment, 150 unique configurations are evaluated for that algorithm, while these are split between the three algorithms when all the algorithms are used. The total of 200 evaluations for all the unique configurations on the minimum, medium, and maximum budget is equivalent to 80 evaluations on the maximum (full) budget, i.e. if we performed 80 evaluations at the maximum number of epochs it would use roughly the same computational power as the 200 BOHB evaluations. In this way, BOHB is able to maximise the specified budget to explore more candidate configurations in the search space.
	
	\begin{table}[!htbp]
		\caption{The number of unique configurations evaluated (No. unique) and the total number of evaluations performed (Total) manually and automatically using BOHB as well as the number of equivalent evaluations on the maximum budget (No. max. equiv.) on the Voltage (V), Methane (M), NYSE (N), and JSE (J) datasets.}
		\label{tab-cash-manual-configuration} 
		\centering
		\resizebox{0.8\columnwidth}{!}{
		\begin{tabular}{lSSSSSSSS}
			\toprule
			
			& \multicolumn{4}{S}{Manual} & \multicolumn{4}{S}{BOHB \{Single/All\}}\\
			\cmidrule(lr{.75em}){2-5} \cmidrule(lr{.75em}){6-9} 
			& {V} & {M} & {N}  & {J} 
			& {V} & {M} & {N}  & {J}\\
			
			\midrule
			{\textit{No. unique}} 			
			& {55} & {48} & {46}  & {46} 
			& {150} & {150} & {150}  & {150}\\
			
% 			\hline
			{\textit{Total}} 			
			& {550} & {480} & {460}  & {460} 
			& {200} & {200} & {200}  & {200}\\
			
% 			\hline
			{\textit{No. max. equiv.}} 
			& {-} & {-} & {-}  & {-} 
			& {80} & {80} & {80}  & {80}\\
			\bottomrule	
		\end{tabular}
			}
	\end{table}
	
	\subsection{Optimal AutoML models across all algorithms }
	In the first experiment we searched the full search space, for different model configurations across all three algorithms. Given the stochastic nature of BOHB, we conducted 10 independent runs of the experiment on all the data sets except for the methane data, where we only conducted 5 runs because of its long runtime.  
	Table~\ref{tab-cash-optimal-model-fraction} shows the number of times each of the candidate models is found by BOHB as the optimal algorithm across these 10 runs as well as the best manual DNN algorithm. The experiment was run 5 times on the methane dataset because on average it took over 41 hours to complete a single run on that dataset since it has 44 different training-validation-test sets (see section~\ref{sec-evaluation}).
	\begin{table}[!htbp]
		\caption{The number of times each of DNN models is selected as optimal by BOHB, and the best manual DNN model (Best M-DNN)}
		\label{tab-cash-optimal-model-fraction} 
		\centering
		\resizebox{0.78\columnwidth}{!}{
			\begin{tabular}{lSSSS}
				\toprule
				& {Voltage} & {Methane} & {NYSE}  & {JSE}\\
				
				\midrule
				{{MLP}} & {\textbf{5}} & {2} & {0}  & {\textbf{9}}\\
				
				% \hline
				{{LSTM}} & {0} & {0} & {\textbf{9}}  & {0}\\
				
				% \hline
				{{CNN}} & {\textbf{5}} & {3} & {1}  & {1}\\
				% \hline
				{{Total runs}} & {10} & {5} & {10}  & {10}\\
				% \hline
				%			{Most optimal model}  & {MLP/CNN} & {-} & {LSTM}  & {MLP}\\
				%			\hline
				{{Best M-DNN}}  & {\textbf{CNN}} & {MLP} & {\textbf{LSTM}}  & {\textbf{MLP/CNN}}\\
				\bottomrule	
			\end{tabular}
		}
	\end{table}
	The optimal model found by BOHB, 9 times out of 10 runs,  is the same as the best manually tuned algorithm for the NYSE and the JSE datasets. It is also interesting to note that BOHB favoured the simpler/faster MLP model over the CNN for the JSE data set.
	On the voltage dataset, the best manually tuned algorithm is found 5 times by BOHB as the optimal algorithm. The other 5 times, it found a different algorithm namely MLP instead of CNN. However, the average performance of the 5 MLP models is better than the average performance of the 5 CNN models (RMSE of 36.21 and RMSE of 36.46). Again the MLP is favoured over the CNN, which shows a bias towards simpler/faster models. The interesting result is the Methane data set where the CNN is favoured over the MLP. However, this is not conclusive given that only 5 runs were performed.
	Regarding the optimal hyperparameters, each run of BOHB finds a different optimal configuration given the same optimal algorithm. However, the variance in the performances of these configurations is low. This is shown by the small standard deviation of the RMSE values achieved by the different configurations (CASH) in the appendix table~\ref{tab:auto-cash-bohb-full}. This confirms the \textit{robustness} of BOHB as suggested by Falkner et al \cite{Falkner2018}. 
	
		\subsection{Effect of increase in budget}
	In the second experiment, we constrained the search space to a single algorithm. For each algorithm, we evaluated 150 unique configurations for a single algorithm instead of 150 configurations across 3 algorithms. This in some sense provided an increase in budget. Practically, searching each algorithm at a time allowed us to set BOHB's maximum budget for each algorithm to the optimal number of epochs found during the manual experiment for that algorithm.  This is a more accurate estimation of the true function to be learnt given that algorithm. However, it resulted in an increase or a decrease of the minimum/maximum budget for certain algorithms on certain datasets, compared to the combined search (see the change in budgets from \textit{All} to \textit{single algorithm} in table~\ref{tab:budgets}). For instance, the maximum budget for the LSTM on the NYSE decreased to 100 from 500, which was chosen for the combined search based on the manually found optimal number of epochs. 
	
	Table~\ref{tab:auto-cash-bohb-full} provides a comparison of the performance of the optimal model found when all algorithms (BOHB-All) are used and when only a single algorithm (BOHB-Single) is used, i.e when the budget - number of configurations evaluated per algorithm - is increased. The mean and deviation of the optimal model from each experiment trained and evaluated over 10 independent runs is shown.
		\begin{table}[!htbp]
	    \centering
		\caption{Comparison of BOHB on all algorithm (BOHB-All) and the best BOHB on the single algorithms (BOHB-Single). Model (M), Slope RMSE (S), Duration RMSE (D), Average slope and duration RMSEs (A)}
		\label{tab:auto-cash-bohb-full}
		\resizebox{0.80\columnwidth}{!}{
			\begin{tabular} {llccc}
				\toprule
				&  & {\textit{BOHB-All}}  & {\textit{BOHB-Single}}& {\textit{\% Delta}}\\
				\midrule
				\multirow{4}{*}{\textit{Voltage}} 
				& {\textit{M}} & {CNN} & {CNN} & {-}\\
				& {\textit{S}} & {$9.70 \pm 0.44$}  & {$\textbf{9.08} \pm \textbf{0.04}$} & {$ \textbf{6.39 }$}\\
				& {\textit{D}} & {$62.97 \pm 0.14$} & {$\textbf{62.35} \pm \textbf{0.02}$} & {$ \textbf{0.98} $}\\
				& {\textit{A}} & {$36.34 \pm 0.29$} & {$\textbf{35.72} \pm \textbf{0.03}$} & {$ \textbf{1.71} $}\\
				\midrule
				\multirow{4}{*}{\textit{Methane}} 
				& {\textit{M}} & {CNN}  & {MLP} & {-}\\
				& {\textit{S}} & {$15.01 \pm 0.88$} & {$\textbf{14.01} \pm \textbf{0.21}$} & {$ \textbf{6.66} $} \\
				& {\textit{D}} & {$47.42 \pm 4.05$} & {$\textbf{40.09} \pm \textbf{6.95}$} & {$ \textbf{15.46}$}\\
				& {\textit{A}} & {$31.22 \pm 2.47$} & {$\textbf{27.05} \pm \textbf{3.58}$} & {$ \textbf{13.36} $}\\
					
				\midrule
				\multirow{4}{*}{\textit{NYSE}} 
				& {\textit{M}} & {LSTM} & {LSTM} & {-}\\
				& {\textit{S}} & {$\textbf{86.61} \pm \textbf{0.03}$} & {$86.62 \pm 0.03$} & {$ {-0.01} $}\\
				& {\textit{D}} & {$\textbf{0.55} \pm \textbf{0.15}$} & {$0.72 \pm 0.30$} & {$ {-30.91} $}\\
				& {\textit{A}} & {$\textbf{43.58} \pm \textbf{0.09}$}  & {$43.67 \pm 0.17$} & {$ {-0.21} $}\\
				
				\midrule
				\multirow{4}{*}{\textit{JSE}} 
				& {\textit{M}} & {CNN}  & {CNN} & {-}\\
				& {\textit{S}} & {$20.00 \pm 0.13$} & {$\textbf{19.96} \pm \textbf{0.17}$} & {$ \textbf{0.20} $}\\
				& {\textit{D}} & {$\textbf{12.46} \pm \textbf{0.18}$} & {$\textbf{12.46} \pm \textbf{0.12}$} & {$ {0.0} $}\\
				& {\textit{A}} & {$16.23 \pm 0.16$} & {$\textbf{16.21} \pm \textbf{0.15}$} & {$\textbf{0.12 }$}\\
				
				\bottomrule
			\end{tabular}
		}
	\end{table}
	By evaluating more configurations per algorithm a much better model is found for the methane data set, but only marginally better model for the voltage and JSE data sets. There is a slight drop in performance for the NYSE. This could be because when the search is constrained to a single algorithm, the best configuration found by BOHB, i.e. LSTM is evaluated with a smaller number of epochs compared to the LSTM found by BOHB when all the algorithms are searched together (100 vs. 500 - see table~\ref{tab:budgets}).
	When BOHB focused on each algorithm, the best model found for each dataset matches the performance of the best manual models. 
	
	\subsection{Summary of the best BOHB models and the manual models}
	We now describe how the performance of the optimal model found by AutoML fares against the model found in the manual selection process. Table \ref{tab-best-automl-manual} shows the trend prediction results of the best algorithm and model selected by BOHB and compares this to the model selected during the manual process. Note that we trained and tested the best AutoML model found by BOHB across 10 independent runs.  The deviation across the 10 runs is also shown to provide an indication of the stability of the model across the runs. 
	\begin{table}[!htbp]
	    \centering
		\caption{Comparison of the best AutoML models and the best manual models. Model (M), Slope RMSE (S), Duration RMSE (D), Average slope and duration RMSEs (A)}
		\label{tab-best-automl-manual}
		\resizebox{0.80\columnwidth}{!}{
			\begin{tabular} {llccc}
				\toprule
				&  & {\textit{BOHB}}  & {\textit{Manual}}& {\textit{\% Delta}}\\
				\midrule
				\multirow{4}{*}{\textit{Voltage}} 
				& {\textit{M}} & {CNN} & {CNN} & {-}\\
				& {\textit{S}} & {$\textbf{9.08} \pm \textbf{0.04}$}  & {$9.24 \pm 0.10$} & {\textbf{1.73}}\\
				& {\textit{D}} & {$\textbf{62.35} \pm \textbf{0.02}$} & {$62.40 \pm 0.13$} & {\textbf{0.08}}\\
				& {\textit{A}} & {$\textbf{35.72} \pm \textbf{0.03}$} & {$35.82 \pm 0.12$} & {\textbf{0.28}}\\
				
				\midrule
				\multirow{4}{*}{\textit{Methane}} 
				& {\textit{M}} & {MLP}  & {MLP} & {-}\\
				& {\textit{S}} & {$\textbf{14.01} \pm \textbf{0.21}$} & {$14.57\pm 0.10$} & {\textbf{3.84}}\\
				& {\textit{D}} & {$\textbf{40.09} \pm \textbf{6.95}$} & {$49.79 \pm 4.85$} & {\textbf{19.48}}\\
				& {\textit{A}} & {$\textbf{27.05} \pm \textbf{3.58}$} & {$32.18 \pm 2.48$} & {\textbf{15.94}}\\
				
				\midrule
				\multirow{4}{*}{\textit{NYSE}} 
				& {\textit{M}} & {LSTM} & {LSTM} & {-}\\
				& {\textit{S}} & {${86.61} \pm {0.03}$} & {$\textbf{86.56} \pm \textbf{0.01}$} & {-0.06}\\
				& {\textit{D}} & {${0.55} \pm {0.15}$} & {$\textbf{0.41} \pm \textbf{0.08}$} & {-34.15}\\
				& {\textit{A}} & {${43.58} \pm {0.09}$}  & {$\textbf{43.49} \pm \textbf{0.05}$} & {-0.21}\\
				
				\midrule
				\multirow{4}{*}{\textit{JSE}} 
				& {\textit{M}} & {CNN}  & {MLP} & {-}\\
				& {\textit{S}} & {${19.96} \pm {0.17}$} & {$\textbf{19.87} \pm \textbf{0.01}$} & {-0.45}\\
				& {\textit{D}} & {$\textbf{12.46} \pm \textbf{0.12}$} & {$12.51 \pm 0.09$} & {\textbf{0.40}}\\
				& {\textit{A}} & {${16.21} \pm {0.15}$} & {$\textbf{16.19} \pm \textbf{0.05}$} & {-0.12}\\
				
				\bottomrule
			\end{tabular}
		}
	\end{table}
For the voltage dataset, the average performance of the best AutoML model showed a marginal increase of 0.28\% on the best manual model, also a CNN, but showed a substantial performance improvement of 15.94\% on the methane data set.There was a marginal drop in performance of 0.21\% and 0.12\% in the AutoML models for the NYSE and JSE datasets respectively. All algorithms identified by AutoML align with the algorithm found in the manual experiments, except for the JSE data set, where the CNN was favoured over the MLP.

In terms of stability of the optimal configuration across 10 runs, the deviation from the average performance is somewhat stable for the voltage, NYSE JSE datasets ($<1\%$), but 1.1\% higher for the methane data set when compared to the stability of the manual models. It must be pointed out that BOHB did not take into account model stability during the search process. Optimal candidate configurations were only evaluated once, while all candidate configurations were evaluated 10 times during the manual process. This could be incorporated into the BOHB experiments, but would substantially increase the running time of the BOHB experiments.

% More discussions need to be added to this section.
%TODO:
% 2. Mention the low risk average risk on NYSE and JSE but high risk on NYSE
% 3. Mention necessity to have a loss metric that captures all the requirements of a specific application

% Table~\ref{tab:cash-runtine} shows the average time (in seconds) taken by BOHB to find the optimal algorithm and its optimal hyperparameter values for each dataset.
% \begin{table}[!htbp]
% 	\caption{Average run time of BOHB for 30 iterations.}
% 	\label{tab:cash-runtine}
% 	\centering
% 	\resizebox{\columnwidth}{!}{
% 		\begin{tabular} {cccc}
% 			\toprule
% 			 {Voltage} & {Methane} & {NYSE} & {JSE} \\
%             \midrule
% 			{$ \pm $} & {$150283.74 \pm 24771.11$} & {$2275.29 \pm 330.70$} & {$1300.14 \pm 144.10$}\\						  
% 			\bottomrule	
% 		\end{tabular}
% 	}
% \end{table}

% Table~\ref{tab:manual-runtine} shows the average runtime of each of the manual model. 
% 	\begin{table}[!htbp]
% 	\caption{Training time (seconds) of each manual model.}
% 	\label{tab:manual-runtine}
% 	\centering
% 	\resizebox{\columnwidth}{!}{
% 		\begin{tabular}{cccc}
% 			\toprule
% 			%			\toprule
% 			 & MLP & LSTM & CNN & \\
% 			\midrule
% 			\textit{Voltage} & {$61.24 \pm 3.69$} & {$17.36 \pm 0.22$} & {$825.00 \pm 6.97$}\\
% 			\midrule
% 			\textit{Methane} & {$3096.43 \pm 342.76$} & {$1113.98 \pm 12.03$} & {$1511.60 \pm 52.93$} \\
% 			\midrule
% 			\textit{NYSE} & {$7.45 \pm 0.33$} & {${0.73} \pm {0.03}$} & {$114.60 \pm 0.41$}\\
% 			\midrule
% 			\textit{JSE} & {$4.57 \pm 0.40$} & {${2.43} \pm {0.15}$} & {$3.64 \pm 0.30$}\\
% 			\bottomrule
% 		\end{tabular}
% 	}
% \end{table}

	\section{Discussion and Conclusions}

    We set out to explore the use of automatic machine learning techniques to automate the algorithm selection and hyperparameter optimisation process for predicting trends in time series data sets. We demonstrated how AutoML tools, specifically the HpBandSter framework, can be effectively used to automatically find an optimal deep neural network configuration. Our BOHB experiments found optimal configurations that produced models that compared well against the average performance and stability levels of configurations found during the manual experiments. Our results show that BOHB is a promising framework for automating algorithm selection and hyperparameter optimisation for trend prediction in time series data sets. 

	Even though trend prediction has broad real world application, there are relatively few studies that apply Deep Neural Networks to this area when compared to applications for computer vision, natural language processing and speech recognition. The results of this study can be used to accelerate research in this area and practical evaluation of these algorithms for real world applications. It is important to note that the proposal in this paper is not for using BOHB as a fully automated solution, but to assist a machine learning practitioner to perform preliminary experiments and to establish baselines for further exploration and tuning.

    The hyperparameter configuration search space identifies key hyperparameter variables and ranges that we found most impacted model performance across all data sets during our manual experiments. We have made our configuration file publicly accessible\footnotemark{} to allow researchers and practitioners to replicate these results and evaluate this approach on other data sets. This could be a first step towards formalising and sharing machine learning knowledge.
    
    \addtocounter{footnote}{-1} %n=6
	\stepcounter{footnote}\footnotetext{\resizebox{!}{0.8\height}{https://github.com/h-kouame/configuration-space-of-auto-cash}}
	
	\subsection{Limitations and future work}
	There are many avenues to probe the results of this work further. Firstly we only tested this on four data sets. While these included all three data sets used in the original TreNet paper \cite{Lin2017}, testing on more data sets is required to probe the generalisation of these findings. The hyperparameter search space was restricted to three vanilla DNN algorithms and can be expanded to include other algorithms and more hyperparameters, but this will of course affect the budget and running time. We used local point data as input features. We could explore the addition of trend lines as input features and add this as an additional variable in the hyperparameter search space.
\bibliographystyle{aaai}

\end{document}